%% file: main.tex
\documentclass[10pt,twocolumn,letterpaper]{article}

\usepackage[pagenumbers]{cvpr} 

\input{preamble}
\definecolor{cvprblue}{rgb}{0.21,0.49,0.74}
\usepackage[pagebackref,breaklinks,colorlinks,allcolors=cvprblue]{hyperref}
\usepackage{graphicx}
\usepackage{booktabs}
\usepackage{pifont}
\usepackage[accsupp]{axessibility}  

\usepackage{adjustbox}
\usepackage{multirow}
\usepackage[table,xcdraw]{xcolor}
\usepackage{epigraph}
\usepackage{soul}
\usepackage[utf8]{inputenc}
\usepackage{tcolorbox}
\usepackage{fontenc}
\usepackage{fontawesome5}

\tcbuselibrary{skins, breakable}
\definecolor{headerblue}{RGB}{74, 107, 214}
\definecolor{boxbg}{RGB}{235, 242, 255}
\definecolor{leftbar}{RGB}{74, 107, 214}
\usepackage{enumitem}

\usepackage{tcolorbox}
\definecolor{opensrcHdr}{HTML}{C8DDFB}
\definecolor{opensrcRow}{HTML}{EAF0FB} 
\definecolor{propriHdr}{HTML}{E0D4F5} 
\definecolor{propriRow}{HTML}{F2EDF9} 
\definecolor{spatialHdr}{HTML}{C8D5F0}
\definecolor{spatialRow}{HTML}{E8EDF8} 

\definecolor{TakeawayPurple}{HTML}{6B4EAD}
\definecolor{LightPurpleBg}{HTML}{F8F7FF}

\newcommand{\modelicon}[1]{\raisebox{-0.15em}{\includegraphics[height=1.0em]{#1}}\hspace{0.1em}}

\def\paperID{00009} 
\def\confName{CVPR}
\def\confYear{2026}

\title{Mind over Space: Can Multimodal Large Language Models Mentally Navigate?}

\author{
Qihui Zhu$^{1}$\thanks{Equal contribution.} \quad
Shouwei Ruan$^{1}$\footnotemark[1] \quad
Xiao Yang$^{2}$ \quad
Hao Jiang$^{3}$ \quad
Yao Huang$^{2,4}$\\
Shiji Zhao$^{1}$ \quad
Hanwei Fan$^{5}$ \quad
Hang Su$^{2}$ \quad
Xingxing Wei$^{1*}$\thanks{Corresponding author.}\\[6pt]
$^{1}$Institute of Artificial Intelligence, Beihang University\\
$^{2}$Dept. of Comp. Sci. and Tech., Institute for AI, Tsinghua-Bosch Joint ML Center,\\
THBI Lab, BNRist Center, Tsinghua University\\
$^{3}$School of Automation Science and Electrical Engineering
, Beihang University\\
$^{4}$college of AI, Tsinghua University\\
$^{5}$Department of Computer Science and Technology , Tsinghua University\\
}

\begin{document}
\maketitle
\input{sec/0_abstract}    
\input{sec/1_intro}
\input{sec/3_benchmark}
\input{sec/4_learning_mental}
\input{sec/5_conclusion}
{
    \small
    \bibliographystyle{ieeenat_fullname}
    \bibliography{main}
}


\end{document}

%% file: sec/0_abstract.tex
\begin{abstract}
Despite the widespread adoption of MLLMs in embodied agents, their capabilities remain largely confined to reactive planning from immediate observations, consistently failing in spatial reasoning across extensive spatiotemporal scales. Cognitive science reveals that Biological Intelligence (BI) thrives on \textbf{\textit{``mental navigation"}}: \textit{the strategic construction of spatial representations from experience and the subsequent mental simulation of paths prior to action}. To bridge the gap between AI and BI, we introduce \textbf{Video2Mental}, a pioneering benchmark for evaluating the mental navigation capabilities of MLLMs. The task requires constructing hierarchical cognitive maps from long egocentric videos and generating landmark-based path plans step by step, with planning accuracy verified through simulator-based physical interaction. Our benchmarking results reveal that mental navigation capability does not naturally emerge from standard pre-training. Frontier MLLMs struggle profoundly with zero-shot structured spatial representation, and their planning accuracy decays precipitously over extended horizons. To overcome this, we propose \textbf{NavMind}, a reasoning model that internalizes mental navigation using explicit, fine-grained cognitive maps as learnable intermediate representations. Through a difficulty-stratified progressive supervised fine-tuning paradigm, NavMind effectively bridges the gap between raw perception and structured planning. Experiments demonstrate that NavMind achieves superior mental navigation capabilities, significantly outperforming frontier commercial and spatial MLLMs.
\end{abstract}

%% file: sec/1_intro.tex
\section{Introduction}
\label{sec:intro}

\begin{figure*}[t]
    \centering
    \includegraphics[width=0.95\textwidth]{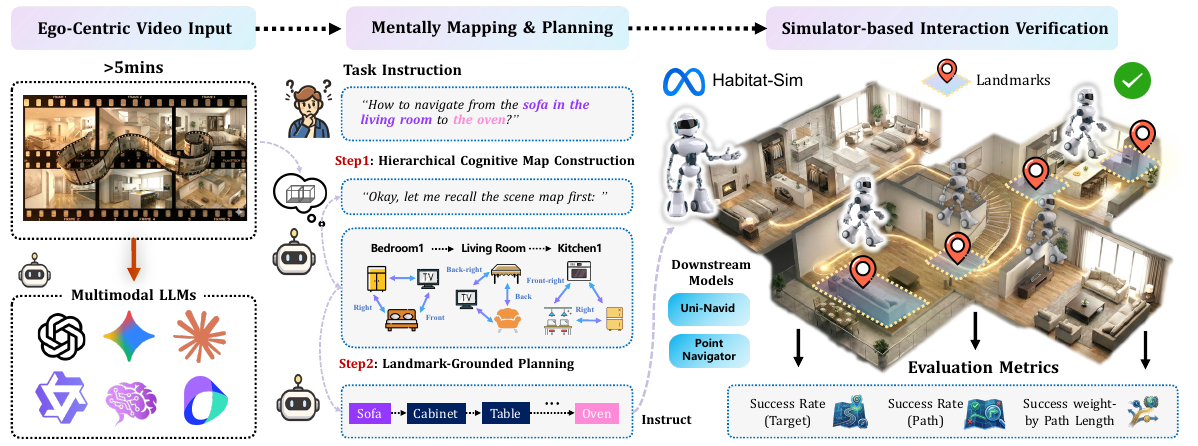}
    \caption{\textbf{Illustration of the Mental Navigation Task for MLLMs.} We define mental navigation as a task requiring MLLMs to comprehend long egocentric videos (over 5 minutes) and perform a two-stage reasoning process. First, the model abstracts a scene cognitive map from the video and outputs it as a structured representation (e.g., JSON). It then infers a landmark-grounded route plan connecting the specified origin and destination. The generated plans are further evaluated in a simulator using downstream navigation expert models across multiple metrics.}
    \label{fig:teaser}
    \vspace{-6mm}
\end{figure*}
The rapid advancement of Multimodal Large Language Models (MLLMs) has equipped embodied agents with strong visual understanding and cross-modal reasoning~\cite{gemini3pro, comanici2025gemini, singh2025openai, bai2025qwen3, openai2023gpt4v_systemcard}, enabling them to map immediate observations to task plans~\cite{ahn2022can, huang2023voxposer, wu2023embodied, zou2025embodiedbrain} and even executable actions~\cite{zitkovich2023rt, driess2023palm, huang2023visual, gu2024conceptgraphs}. Despite this surface-level competence, a critical bottleneck persists: current deployments of MLLMs in embodied scenarios are almost entirely governed by short-horizon, local \textit{reactive planning}~\cite{team2025robobrain, majumdar2024openeqa, cai2024bridging, qi2025vln, cheng2024navila, zhang2024uni}. Although some recent works have attempted to incorporate longer temporal history to assist planning via visual episodic memory or token pruning, recent studies reveal that as the spatio-temporal horizon of a task expands, the spatial reasoning performance of frontier MLLMs still suffers a precipitous decline~\cite{yang2025thinking, lin2025mmsi, yang2025cambrian}. The root cause of this degradation is that existing models struggle to maintain long-range spatial dependencies. More fundamentally, they cannot infer global environment layouts from streaming egocentric video; \textit{i.e.}, they lack the capacity to construct \textit{spatial mental representations}~\cite{yang2025thinking, yin2025spatial}. Consequently, when confronted with long-horizon spatial reasoning tasks, such as deriving a navigation plan from extended video observations, these frontier models are unable to transcend the dual barriers of constrained spatial memory and inaccurate spatial representation.

Research in cognitive science offers a key insight into resolving this impasse: \textit{Biological Intelligence} (BI) in complex environments depends profoundly on an innate capacity for \textbf{mental navigation}. Unlike current artificial embodied navigation systems~\cite{zhang2024uni, cheng2024navila}, which rely on step-wise policy or action planning from immediate visual observations, biological agents can strategically abstract and construct structured spatial representation, which known as \textit{cognitive maps}~\cite{o1978hippocampus, schiller2015memory}, from past exploratory experience alone, and simulate prospective paths internally before executing any physical action~\cite{neupane2024mental, bellmund2018navigating, eichenbaum2017role, broadbent2004spatial}. This mechanism liberates biological intelligence from absolute dependence on real-time perception, enabling proactive, global planning across extended spatio-temporal scales.

To bridge the gap between current MLLMs and biological intelligence in long-horizon spatial reasoning and planning, this work addresses a central question:
\textit{How can we endow MLLMs with \textbf{biological-like mental navigation}, enabling them to internalize spatial representations from streaming egocentric observations and accomplish long-horizon navigation planning?}

\noindent Toward this goal, we make the \textbf{following contributions}:

\ding{182} \textbf{A novel mental navigation task and evaluation benchmark for MLLMs.} To systematically quantify the capability boundaries of existing MLLMs in long-horizon spatial reasoning, we formally define the \textit{mental navigation} task. As illustrated in Fig.~\ref{fig:teaser}, given egocentric videos exceeding five minutes, an MLLM must deduce a landmark-based global path connecting a specified origin and destination. To explicitly probe the model’s internal spatial representations, it is required to first generate a textual \textit{cognitive map} satisfying strict topological constraints before producing the navigation plan. This design removes reliance on local perception and directly evaluates the model’s ability to integrate spatial memory across viewpoints and perform internal path simulation. Unlike prior spatial reasoning benchmarks that rely on simplified protocols such as multiple-choice questions or numeric matching, we validate generated plans through physical interaction in the Habitat simulator, ensuring faithful evaluation of their physical correctness.

\begin{figure*}[t]
    \centering
    \includegraphics[width=0.85\textwidth]{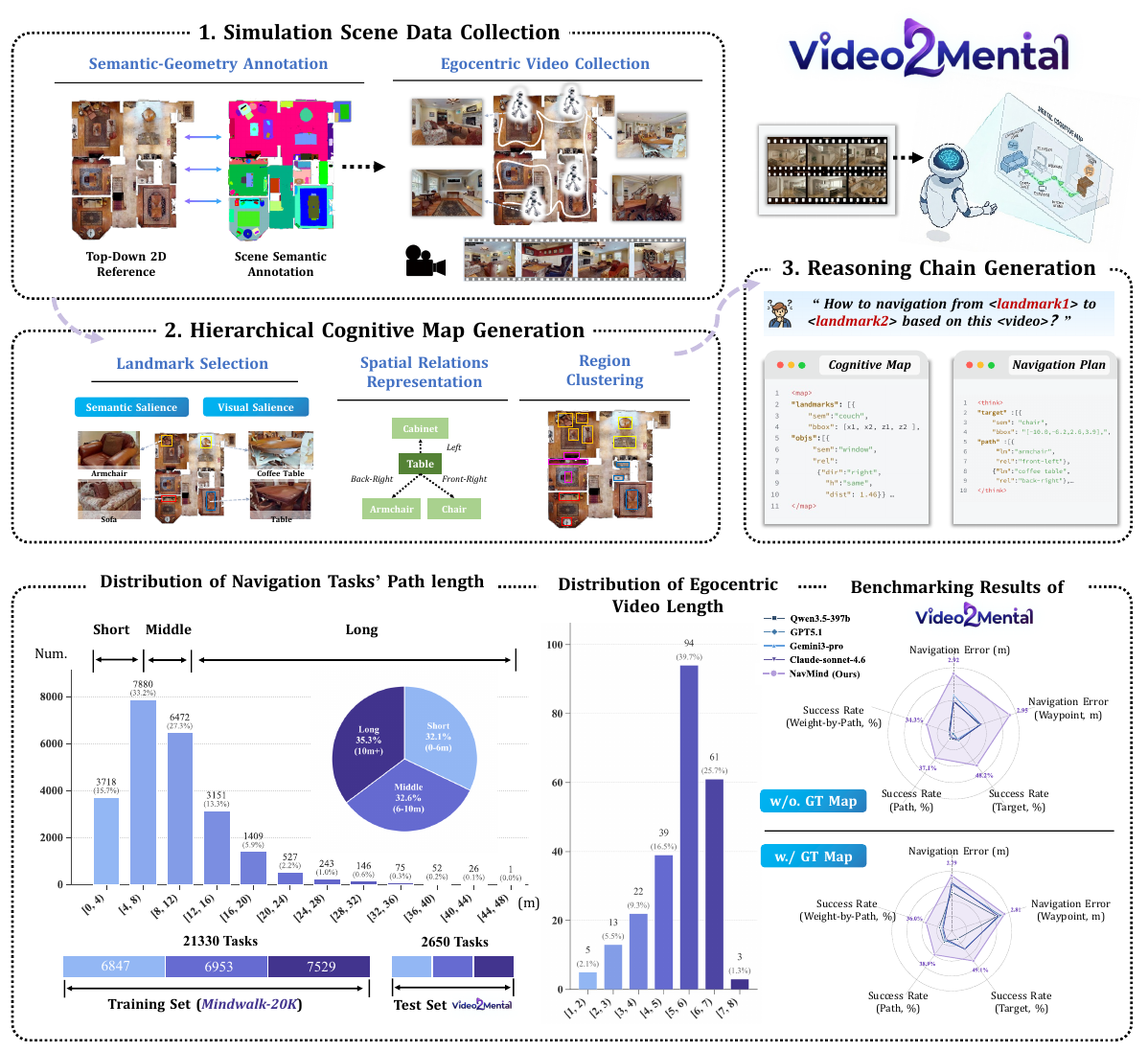}
    \caption{\textbf{Data generation pipeline and benchmark analysis.} Video2Mental Benchmark Construction and Statistics. We collect simulator-based semantic annotations and egocentric exploration videos, synthesize hierarchical cognitive maps via rule-based landmark selection, and extract landmark-grounded path planning sequences. The resulting dataset contains over \textbf{24k mental navigation tasks} with average path lengths exceeding 9 meters and more than 80\% of videos longer than 4 minutes. Preliminary evaluations show that existing MLLMs struggle to demonstrate genuine mental navigation capability.}
    \label{fig:bench}
    \vspace{-0.4cm}
\end{figure*}

Building on this formulation, we construct \textbf{\textit{Video2Mental}}, a large-scale benchmark comprising 23,700 high-difficulty mental navigation samples, with a dedicated test split of 2,300 samples. As shown in Fig.~\ref{fig:bench}, the benchmark is organized into three difficulty levels based on spatio-temporal span and evaluated using multi-dimensional metrics, including cognitive map accuracy and simulator-based navigation success. Extensive evaluation of frontier MLLMs reveals two sobering insights: \textit{\textbf{1)}} in stark contrast to human spatial cognition, mental navigation is not a capability that \textit{naturally emerges} from large-scale vision-language pre-training; \textit{\textbf{2)}} even when ground-truth cognitive maps are supplied as input, models still produce severe planning errors. This conclusively demonstrates that the bottleneck is deeply rooted in the absence of structured spatial reasoning rather than perceptual deficiencies alone.

\ding{183} \textbf{An spatial reasoning model with mental navigation capability.} The insights above point to a clear path forward: explicitly teaching models to construct and operate over structured representations from long-horizon video data is the key to bridging the AI--BI divide. We therefore propose \textbf{\textit{NavMind}}, a reasoning model that \textit{internalizes} mental navigation as a structured reasoning capability. Rather than targeting step-wise reactive planning, NavMind employs explicit, fine-grained cognitive maps as intermediate learnable representations to support global navigation planning. Built upon the Qwen3-VL~\cite{bai2025qwen3} architecture, NavMind is trained on the training split of Video2Mental through a two-stage process. To equip the model with deep long-horizon spatial reasoning, we propose a \textbf{difficulty-stratified progressive} Supervised Fine-Tuning (SFT) paradigm. By employing rejection sampling to filter out low-perplexity, simplistic trajectories, we steer the optimization toward difficult samples that demand deep spatial reasoning rather than mere pattern memorization. Evaluations confirm that NavMind acquires robust mental navigation capabilities, when deployed as a reusable planning module for VLN agents, it provides stable global planning signals that yield consistent performance improvements across environments.

We envision that Video2Mental will catalyze progress in long-horizon spatial reasoning and promote evaluation through physical interaction rather than superficial response pattern matching in MLLM-based embodied agents. Furthermore, NavMind provides a highly effective and robust baseline for the mental navigation task, \textit{paving the way for future exploration into brain-inspired cognitive architectures for embodied AI.}

%% file: sec/3_benchmark.tex
\section{Mental Navigation: Task Formulation}
\begin{table*}[t]
\centering
\caption{\textbf{Evaluation results on the Video2Mental benchmark (Part 1).} All MLLMs are required to first generate scene cognitive maps from egocentric videos and then infer landmark-grounded route plans to accomplish mental navigation task. We report both the text-based static evaluation metrics ($\text{NE}$ / $\text{SR}_{\text{t}}$) and the simulator-based interactive validation metrics ($\text{SR}_{\text{p}}$ / $\text{SPL}$).}
\label{table:benchmark_results_think}
\renewcommand{\arraystretch}{1.1}
\setlength{\tabcolsep}{3.6pt}
\small
\begin{adjustbox}{max width=\textwidth}
\begin{tabular}{@{}lc cccc cccc cccc cccc@{}}
\toprule
\multirow{2}{*}{\textbf{Models}} & \multirow{2}{*}{\textbf{Rank}} & \multicolumn{4}{c}{\textbf{Overall}} & \multicolumn{4}{c}{\textbf{Short}} & \multicolumn{4}{c}{\textbf{Middle}} & \multicolumn{4}{c}{\textbf{Long}} \\
\cmidrule(lr){3-6} \cmidrule(lr){7-10} \cmidrule(lr){11-14} \cmidrule(lr){15-18}
 & & {\scriptsize\textbf{NE}}{\scriptsize$\downarrow$} & {\scriptsize$\textbf{SR}_{\textbf{t}}$}{\scriptsize$\uparrow$} & {\scriptsize$\textbf{SR}_{\textbf{p}}$}{\scriptsize$\uparrow$} & {\scriptsize\textbf{SPL}}{\scriptsize$\uparrow$}
 & {\scriptsize\textbf{NE}}{\scriptsize$\downarrow$} & {\scriptsize$\textbf{SR}_{\textbf{t}}$}{\scriptsize$\uparrow$} & {\scriptsize$\textbf{SR}_{\textbf{p}}$}{\scriptsize$\uparrow$} & {\scriptsize\textbf{SPL}}{\scriptsize$\uparrow$}
 & {\scriptsize\textbf{NE}}{\scriptsize$\downarrow$} & {\scriptsize$\textbf{SR}_{\textbf{t}}$}{\scriptsize$\uparrow$} & {\scriptsize$\textbf{SR}_{\textbf{p}}$}{\scriptsize$\uparrow$} & {\scriptsize\textbf{SPL}}{\scriptsize$\uparrow$}
 & {\scriptsize\textbf{NE}}{\scriptsize$\downarrow$} & {\scriptsize$\textbf{SR}_{\textbf{t}}$}{\scriptsize$\uparrow$} & {\scriptsize$\textbf{SR}_{\textbf{p}}$}{\scriptsize$\uparrow$} & {\scriptsize\textbf{SPL}}{\scriptsize$\uparrow$} \\
\midrule
\rowcolor{propriHdr}
\multicolumn{18}{c}{\textbf{\textit{(A) Open-Source Multimodal Large Language Models}}} \\
\modelicon{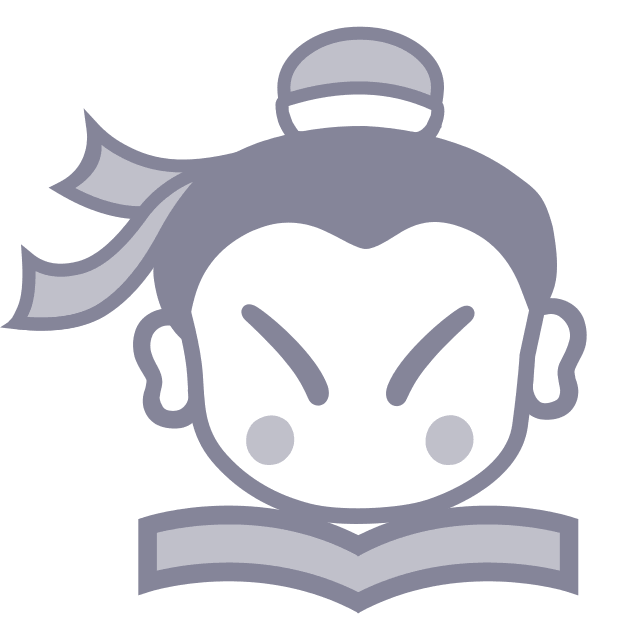}~InternVL-3.5-8B & 10 &6.64	&0.9	&1.6 &1.4	&6.64	&1.0	&1.9	&1.9	&6.24	&1.6	&1.6	&1.3	&7.07	&0.5	&1.2	&1.0 \\
\modelicon{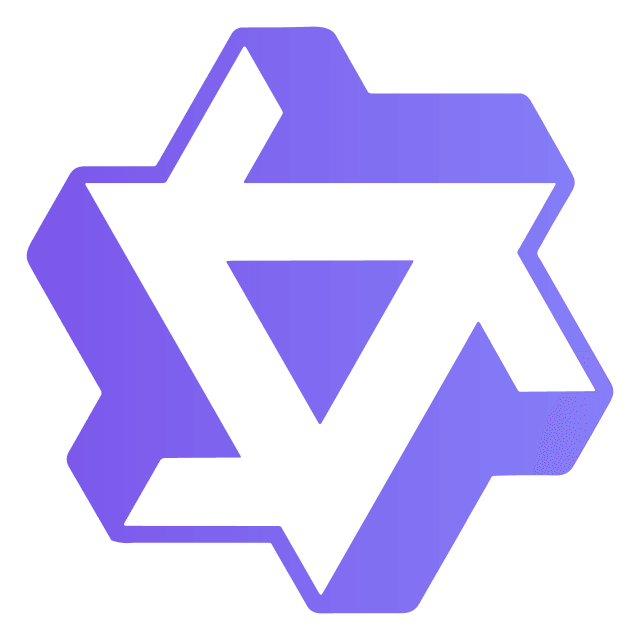}~Qwen3-VL-8B & 7 & 6.45 & 4.4 & 4.4 & 4.0 & 5.20 & 6.1 & 7.0 & 6.1 & 5.85 & 3.7 & 3.5 & 3.2 & 8.18 & 3.5 & 2.9 & 2.7 \\
\modelicon{icons/qwen.png}~Qwen3-VL-30B & 8 & 6.90 & 4.3 & 5.2 & 4.4 & 5.74 & 4.2 & 5.8 & 4.6 & 6.12 & 5.7 & 4.2 & 3.5 & 8.64 & 3.2 & 5.5 & 5.1 \\
\modelicon{icons/qwen.png}~Qwen3.5-397B & \cellcolor{green!5}3 & 5.19 & 10.7 & 7.4 & 5.9 & 4.60 & 13.6 & 7.2 & 5.3 & 4.32 & 10.3 & 7.4 & 6.0 & 6.67 & 9.2 & 7.6 & 6.3 \\
\midrule
\rowcolor{opensrcHdr}
\multicolumn{18}{c}{\textbf{\textit{(B) Proprietary Multimodal Large Language Models}}} \\
\modelicon{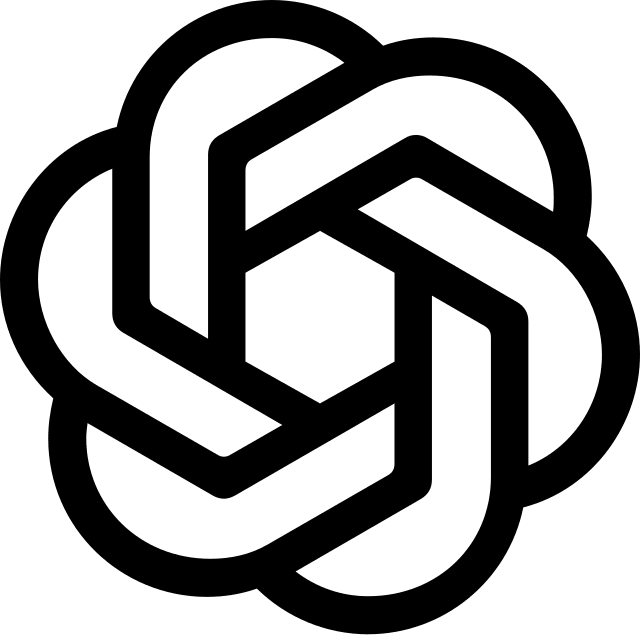}~GPT-5.1 & \cellcolor{green!2}4 & 5.28 & 10.3 & 0.6 & 5.2 & 4.31 & 13.4 & 9.2 & 6.8 & 4.72 & 11.7 & 6.3 & 5.2 & 6.66 & 6.2 & 4.4 & 3.9 \\
\modelicon{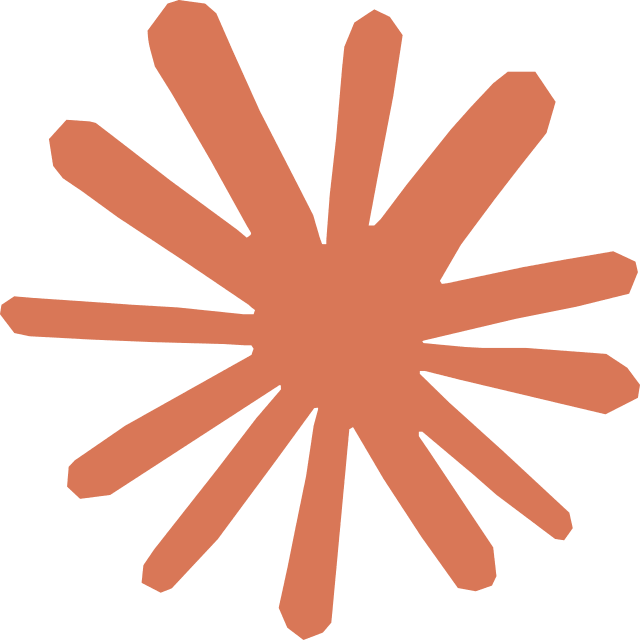}~Claude-Sonnet-4.6 & 5 & 5.28 & 9.8 & 5.8 & 4.7 & 4.45 & 11.1 & 7.9 & 6.6 & 4.19 & 13.7 & 3.6 & 2.5 & 7.17 & 4.4 & 5.9 & 5.0 \\
\modelicon{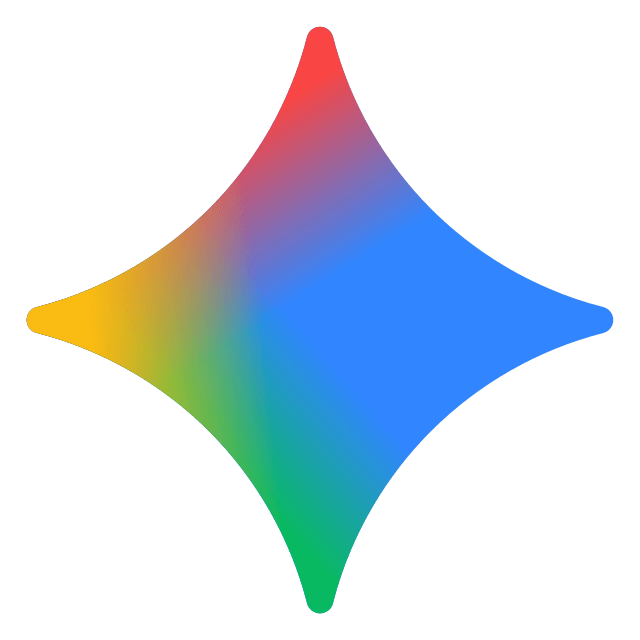}~Gemini-3-Pro & 6 & 4.73 & 8.3 & 3.9 & 3.2 & 4.65 & 8.7 & 4.2 & 3.2 & 3.97 & 8.6 & 4.1 & 3.5 & 6.47 & 7.6 & 3.5 & 2.9 \\
\midrule
\rowcolor{spatialHdr}
\multicolumn{18}{c}{\textbf{\textit{(C) Spatial Reasoning Models}}} \\
\modelicon{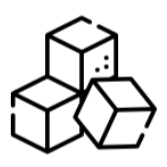}~Cambrian-S & 9 & 9.70 & 1.2 & 4.9 & 4.3 & 9.34 & 2.1 & 3.9 & 3.6 & 10.04 & 0.3 & 4.1 & 3.7 & 9.72 & 1.3 & 5.4 & 4.6 \\
\modelicon{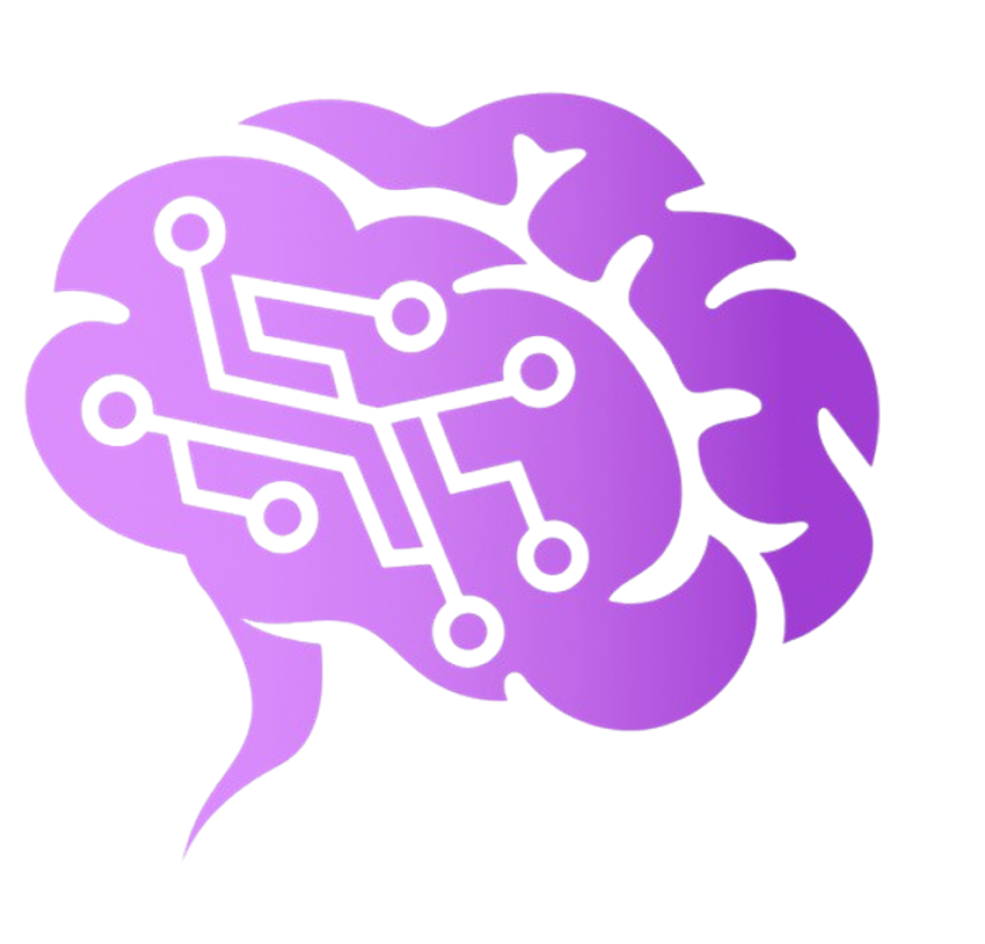}~RynnBrain-8B & 11 & 6.13 & 0.0 & 0.0 & 0.0 & 8.43 & 0.0 & 0.0 & 0.0 & 3.75 & 0.0 & 0.0 & 0.0 & 8.75 & 0.0 & 0.0 & 0.0 \\
\midrule
\rowcolor{propriHdr!35}\modelicon{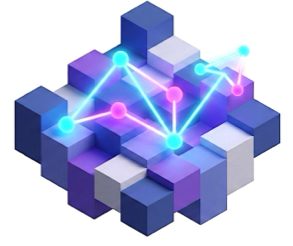}~NavMind-Stage1 (Ours) & \cellcolor{green!10}2 & 2.92 & 48.2 & 37.1 & 34.3 & 2.39 & 51.1 & 40.4 & 36.8 & 2.51 & 50.9 & 38.6 & 36.0 & 3.78 & 43.1 & 32.7 & 30.6 \\
\rowcolor{propriHdr!35}\modelicon{icons/navmind.png}~NavMind-Stage2 (Ours) & \cellcolor{green!20}1 & 2.92 & 48.8 & 38.0 & 35.2 & 2.44 & 50.3 & 40.1 & 36.5 & 2.45 & 53.1 & 39.9 & 36.8 & 3.77 & 43.6 & 34.5 & 32.7 \\
\bottomrule
\end{tabular}
\end{adjustbox}
\end{table*}
To evaluate the capability boundaries of MLLMs in long-horizon spatial reasoning, we formally introduce the \textbf{\textit{Mental Navigation (MN)}} task. 
Inspired by cognitive science, biological agents typically construct internal spatial representations (e.g., hippocampal cognitive maps~\cite{o1978hippocampus}) to integrate egocentric observations and mentally simulate routes before executing navigation behaviors~\cite{bellmund2018navigating, andreano2009sex}. Following this principle, MN requires an MLLM to first infer a structured spatial representation from visual observations and then perform landmark-grounded route planning without real-time environmental feedback.
As illustrated in Fig.~\ref{fig:teaser}, a mental navigation instance is defined as $\mathcal{I} = (V, q)$. The perceptual input is an egocentric video sequence $V = \{f_1, \ldots, f_T\}$ with associated camera poses $(x_i, y_i, z_i, \theta_{\mathrm{yaw_i}})$. The task objective is specified by a natural language query $q=(s_{\mathrm{src}}, s_{\mathrm{tgt}})$ describing the start and target locations. 
Unlike conventional end-to-end VLN tasks that directly output primitive control actions, MN requires the model to generate a structured output consisting of a hierarchical cognitive map $\mathcal{M}$ with a navigation reasoning chain $\mathcal{W}$. The reasoning chain defines a landmark-grounded plan $P$ connecting the conceptual start $(s_{\mathrm{src}}, bbox_{\mathrm{src}})$ and goal $(s_{\mathrm{tgt}}, bbox_{\mathrm{tgt}})$ within the physical scene. Specifically, the plan is represented as an ordered sequence of cognitive steps $P_i=(lm_i, sem_{i}, rel_{i}, bbox_{i})$, where each step corresponds to a spatial landmark $lm_i$, specifying its semantic label $sem_{i}$, spatial bounding box $bbox_{i}$, and the expected egocentric spatial relation $rel_{i}$ between the agent and the landmark.

\section{Video2Mental Benchmark}

\subsection{Overview}

To systematically evaluate and improve MLLMs' mental navigation capabilities, we introduce the Video2Mental dataset. It comprises ~24k step-by-step annotated samples collected from 246 high-fidelity Habitat-Sim~\cite{savva2019habitat, szot2021habitat} indoor scenes (HM3D~\cite{ramakrishnan2021habitat} and MP3D~\cite{chang2017matterport3d}). Each sample provides a strictly aligned quartet: \textbf{\textit{i)}} A semantic geometry substrate, \textbf{\textit{ii)}} an egocentric exploratory video with continuous pose tracking, \textbf{\textit{iii)}} a hierarchical cognitive map, and \textbf{\textit{iv)}} a landmark-grounded navigation reasoning chain paired with its shortest-path trajectory.

As shown in Fig.~\ref{fig:teaser}, tasks are stratified by path length into Short (0–6m), Medium (0–10m), and Long (10–48m). The dataset is split into 21,330 training and 2,650 testing instances. To rigorously assess generalization, the test set is further divided into seen and unseen environments.

\subsection{Dataset Construction}

\noindent \textbf{Spatial reference.} We first establish a unified top-down 2D reference frame by rendering an orthographic floorplan view. From the simulator’s semantic annotations, we extract all object instances with their semantic categories and 3D axis-aligned bounding boxes, and a world-to-floorplan pixel mapping is recorded to ensure consistency across subsequent map construction and visualization.

\noindent \textbf{Egocentric video generation.} We generate egocentric videos by having the agent perform random tours within the navigable floor space. At each frame, we log the agent’s 3D position and yaw $\left(x_i, y_i, z_i, \theta_{\text {yaw}_i}\right)$. The agent follows shortest-path navigation to visit globally sampled waypoints, performing a 360° scan at each before proceeding. Videos are recorded at 640$\times$480 resolution.

\noindent \textbf{Hierarchical cognitive map representation and generation.} Mental navigation fundamentally relies on the ability to abstract and maintain a global spatial memory. We represent the cognitive map as a hierarchical structure $\mathcal{M}=(\mathcal{R}, \mathcal{L}, \mathcal{O})$ which consists of three levels: Region($\mathcal{R}$) → Landmark ($\mathcal{L}$) →Object($\mathcal{O}$) . This aligns with the multi-scale spatial encoding found in the biological hippocampus and entorhinal cortex.


\textbf{\textit{1)}} Selecting semantically and visually salient landmarks.
We first identify semantically and visually salient landmarks to serve as structural anchors of the cognitive map. Low-information background elements(\eg, floors, carpets, \etc) are removed, and the remaining objects are ranked by horizontal footprint area to select stable, visually prominent landmarks  (\eg, sofa, bed, tables, \etc). This step reduces hundreds of atomic objects into a compact set of interpretable navigation landmarks. Each landmark 
 $lm \in \mathcal{L}$ is associated with a semantic label $sem_{lm}$ and localized using a 2D world-coordinate bounding box
 $bbox_{lm}=[x_{min}, x_{max}, z_{min}, z_{max}]$, which serves as a spatial waypoint for navigation reasoning.

\textbf{\textit{2)}} Modeling object–landmark spatial relations. 
Non-landmark objects are hierarchically linked to their nearest anchor landmark using an egocentric spatial descriptor $(dir, h, dist)$. Here, $dir$ represents one of eight discrete bearings (45° bins), $h \in \{same, on\}$ encodes vertical relations, and $dist$ denotes the Euclidean distance. A coarse-to-fine assignment strategy prioritizes nearby objects within the same room while expanding the search radius in sparse layouts, ensuring both precision and coverage of local spatial structure.

\textbf{\textit{3)}} Building region-level spatial structure. 
To capture higher-level spatial organization, landmarks are grouped into regions using affinity-based spectral clustering. Pairwise landmark distances are converted into an affinity matrix using a Gaussian kernel: 
$A_{i j}=\exp \left(-\frac{d_{i j}^2}{2 \sigma^2}\right)$
where the bandwidth $\sigma$ is set to the median of non-zero distances to adapt to scene density. Spectral clustering then partitions landmarks into region clusters representing functional spatial zones (e.g., living rooms or bedrooms).

To ensure the absolute consistency of the world-centered representation, we strictly enforce a right-handed coordinate system: $+X$ aligns with the global right, $+Y$ represents the vertical up-axis, and $+Z$ denotes the global forward. Spatial azimuths are quantified following standard global bearing conventions: 0° corresponds to $+Z$, 90° to $+X$, 180° to $-Z$, and 270° to $-X$. This coordinate foundation compels the MLLMs to perform genuine spatial transformations rather than relying on superficial 2D pixel-level heuristics.

\noindent \textbf{Landmark-grounded reasoning chain construction.} Given the cognitive map, we generate executable navigation trajectories by sampling start–goal entities, selecting reachable poses, and computing shortest paths with the simulator planner, after which the paths are discretized. Based on these trajectories, we construct landmark-grounded reasoning chains as supervision signals, compressing long routes into a small set of human-readable intermediate landmark cues. To ensure consistency and coverage, intermediate cues are inserted only for long path segments that are not grounded by landmarks, while redundant adjacent cues and short steps are merged to keep the reasoning chain concise and compact.

\subsection{Evaluation Protocol}
We partition our metrics into two distinct evaluation tracks: \textbf{\textit{1)}} Text-based Static Evaluation and \textbf{\textit{2)}} Simulator-based Interactive Evaluation. Let the ground-truth target position be $p^\star$ and the predicted target position be $\hat{p}$.

\noindent\textbf{\textit{ \sethlcolor{Lavender!45}\hl{1) Text-based Static Evaluation.}}} Evaluate the MLLMs' responds without relying on environmental feedback. 

\noindent\ding{182}~\textbf{Landmark-Mean IoU:} We match predicted landmark coordinate boxes to ground-truth under semantic and overlap constraints, and average IoU over all ground-truth landmarks to quantify geometric alignment and scale accuracy.

\noindent\ding{183}~\textbf{Landmark-F1:} Evaluates the semantic and spatial completeness of the map. We perform landmark-level semantic and spatial set matching, then compute precision/recall and F1 to measure how well key landmark anchors are recovered.

\noindent\ding{184}~\textbf{NE (Navigation Error):} Calculated as $\|\hat{p} - p^\star\|$, representing the Euclidean distance between the predicted and ground-truth target position.

\noindent\ding{185}~\textbf{$\text{NE}_{\text{waypoint}}$:} Convert the $(lm, rel)$ sequence into global waypoints, and compute the distance between the final waypoint and the ground-truth target, capturing end-to-end deviation from reasoning to an executable trajectory.

\noindent\ding{186}~\textbf{$\text{SR}_{\text{t}}$ (Target Success Rate):} A binary indicator of target localization success, defined as $\text{NE} < 1\text{m}$, reflecting whether the planned target successfully aligns with the destination.

\noindent\textbf{\textit{ \sethlcolor{Periwinkle!25}\hl{2) Simulator-based Interactive Evaluation.}}} Deploys the MLLM's generated plans within the Habitat-Sim~\cite{savva2019habitat, szot2021habitat} pointnavigator to strictly evaluate their physical executability and efficiency.

\noindent\ding{182}~\textbf{$\text{SR}_{\text{p}}$ (Execution-verified Path Success Rate):} A rigorous executability metric. We convert the thought chain into waypoints and register a success only if the path is physically executable within the environment and NE $< 1\text{m}$.

\noindent\ding{183}~\textbf{SPL (Success weighted by Path Length):} Evaluates navigation efficiency by weighting the execution success against the path length, penalizing detours relative to the theoretical shortest path.

\begin{table*}[t]
\centering
\caption{\textbf{Evaluation results on the Video2Mental benchmark (Part 2).} We further explore the upper-bound mental navigation capabilities of MLLMs when explicitly provided with the ground-truth cognitive map as the reasoning context.}
\label{table:benchmark_results_think_mapgt}
\renewcommand{\arraystretch}{1.1}
\setlength{\tabcolsep}{3.6pt}
\small
\begin{adjustbox}{max width=\textwidth}
\begin{tabular}{@{}lc cccc cccc cccc cccc@{}}
\toprule
\multirow{2}{*}{\textbf{Models}} & \multirow{2}{*}{\textbf{Rank}} & \multicolumn{4}{c}{\textbf{Overall}} & \multicolumn{4}{c}{\textbf{Short}} & \multicolumn{4}{c}{\textbf{Middle}} & \multicolumn{4}{c}{\textbf{Long}} \\
\cmidrule(lr){3-6} \cmidrule(lr){7-10} \cmidrule(lr){11-14} \cmidrule(lr){15-18}
 & & {\scriptsize\textbf{NE}}{\scriptsize$\downarrow$} & {\scriptsize$\textbf{SR}_{\textbf{t}}$}{\scriptsize$\uparrow$} & {\scriptsize$\textbf{SR}_{\textbf{p}}$}{\scriptsize$\uparrow$} & {\scriptsize\textbf{SPL}}{\scriptsize$\uparrow$}
 & {\scriptsize\textbf{NE}}{\scriptsize$\downarrow$} & {\scriptsize$\textbf{SR}_{\textbf{t}}$}{\scriptsize$\uparrow$} & {\scriptsize$\textbf{SR}_{\textbf{p}}$}{\scriptsize$\uparrow$} & {\scriptsize\textbf{SPL}}{\scriptsize$\uparrow$}
 & {\scriptsize\textbf{NE}}{\scriptsize$\downarrow$} & {\scriptsize$\textbf{SR}_{\textbf{t}}$}{\scriptsize$\uparrow$} & {\scriptsize$\textbf{SR}_{\textbf{p}}$}{\scriptsize$\uparrow$} & {\scriptsize\textbf{SPL}}{\scriptsize$\uparrow$}
 & {\scriptsize\textbf{NE}}{\scriptsize$\downarrow$} & {\scriptsize$\textbf{SR}_{\textbf{t}}$}{\scriptsize$\uparrow$} & {\scriptsize$\textbf{SR}_{\textbf{p}}$}{\scriptsize$\uparrow$} & {\scriptsize\textbf{SPL}}{\scriptsize$\uparrow$} \\
\midrule
\rowcolor{propriHdr}
\multicolumn{18}{c}{\textbf{\textit{(A) Open-Source Multimodal Large Language Models}}} \\
\modelicon{icons/internlm.png}~InternVL-3.5-8B & 7 &5.92 &11.6 &10.3 &8.9 & 6.31 &7.7 &7.7 &7.7 &4.31 &15.6 &12.5 &10.7 &7.10 &10.8 &10.8 &8.4 \\
\modelicon{icons/qwen.png}~Qwen3-VL-8B & 8 & 6.15 & 11.5 & 13.4 & 12.4 & 5.19 & 13.3 & 15.8 & 14.5 & 5.56 & 13.5 & 14.4 & 13.2 & 7.50 & 8.1 & 10.2 & 9.8 \\
\modelicon{icons/qwen.png}~Qwen3-VL-30B & 5 & 4.26 & 26.6 & 13.5 & 12.4 & 3.16 & 29.9 & 17.6 & 16.2 & 3.68 & 30.0 & 11.2 & 10.1 & 6.09 & 19.7 & 11.0 & 10.0 \\
\modelicon{icons/qwen.png}~Qwen3.5-397B & 6 & 4.36 & 12.8 & 10.1 & 9.6 & 2.52 & 16.7 & 13.5 & 13.0 & 3.57 & 13.7 & 7.2 & 6.7 & 6.17 & 8.1 & 9.6 & 9.1 \\
\midrule
\rowcolor{opensrcHdr}
\multicolumn{18}{c}{\textbf{\textit{(B) Proprietary Multimodal Large Language Models}}} \\
\modelicon{icons/openai.png}~GPT-5.1 & \cellcolor{green!2}4 & 3.61 & 29.3 & 15.3 & 13.3 & 2.51 & 34.0 & 16.7 & 15.1 & 3.24 & 31.6 & 15.1 & 12.9 & 4.95 & 23.7 & 14.2 & 12.0 \\
\modelicon{icons/claude.png}~Claude-Sonnet-4.6 & \cellcolor{green!10}2 & 3.50 & 30.0 & 13.0 & 11.9 & 2.37 & 40.0 & 8.7 & 8.0 & 3.15 & 26.7 & 15.1 & 13.9 & 4.98 & 23.5 & 14.8 & 13.6 \\
\modelicon{icons/gemini.png}~Gemini-3-Pro & \cellcolor{green!5}3 & 3.45 & 29.6 & 13.9 & 12.2 & 2.54 & 32.4 & 17.6 & 15.6 & 3.04 & 33.0 & 11.7 & 10.1 & 4.74 & 24.0 & 10.8 & 9.5 \\
\midrule
\rowcolor{spatialHdr}
\multicolumn{18}{c}{\textbf{\textit{(C) Spatial Reasoning Models}}} \\
\modelicon{icons/cambrian.png}~Cambrian-S & 10 & 7.81 & 3.3 & 8.7 & 7.0 & 7.65 & 3.4 & 8.7 & 8.0 & 7.90 & 3.3 & 9.4 & 8.1 & 7.89 & 3.2 & 8.1 & 7.0 \\

\modelicon{icons/rynn.png}~RynnBrain-8B & 9 & 8.57 & 8.3 & 8.3 & 4.5 & 6.89 & 12.5 & 12.4 & 9.9 & 6.13 & 12.1 & 11.5 & 5.0 & 10.62 & 0.1 & 0.0 & 0.0 \\
\midrule
\rowcolor{propriHdr!35}\modelicon{icons/navmind.png}~NavMind-GTMap (Ours) & \cellcolor{green!20}1 & 2.79 & 49.1 & 38.9 & 36.0 & 2.33 & 52.1 & 41.4 & 37.6 & 2.24 & 53.6 & 42.7 & 39.2 & 3.70 & 42.4 & 33.5 & 31.6 \\
\bottomrule
\end{tabular}
\end{adjustbox}
\end{table*}

\section{What Limits Mental Navigation Capability in MLLMs}

Tab.~\ref{table:benchmark_results_think} and Tab.~\ref{table:benchmark_results_think_mapgt} summarize the performance of representative MLLMs on the proposed Video2Mental benchmark. To isolate the primary failure modes, we utilize two distinct settings: Mental Navigation (MN), where the model predicts both the cognitive map $\mathcal{M}$ and navigation plan $\mathcal{W}$ given $(V,q)$; and MN (w/ GT-Map), an oracle-guided setting where the ground-truth $\mathcal{M}$ is provided to isolate reasoning from perceptual noise. Our investigation yields three key insights.

\subsection{The Emergence Gap in Mental Navigation.}
\label{sec:Emergence Gap}
A comprehensive evaluation under the MN setting reveals that current MLLMs exhibit a profound "emergence gap" in mental navigation. Despite their proficiency in reactive embodied tasks, these models almost entirely fail at mental navigation. As shown in Tab.~\ref{table:benchmark_results_think}, the average $SR_{\textbf{t}}$ and $SR_{\textbf{p}}$ remain as low as 5.54\% and 3.76\%, respectively. Even frontier closed-source models and the latest Qwen3.5 fall significantly short of practical utility. Diagnostic analysis of the predicted cognitive maps shows a Landmark-Mean IoU below 5\% and Landmark-F1 under 35\%. These results confirm that mental navigation does not naturally emerge from standard vision-language pre-training. We attribute this to two systemic factors: \textbf{\textit{1)}} the absence of large-scale, reasoning-centric data for long-horizon spatial integration, and \textbf{\textit{2)}} the prevailing end-to-end paradigm, which opaquely entangles spatial memory and action planning within a single autoregressive process, preventing the model from forming stable, independent world models.

\subsection{Spatial Reasoning, Not Perception, is the Primary Bottleneck.}

Following the catastrophic failures observed in Sec.~\ref{sec:Emergence Gap}, we investigate whether the bottleneck resides in visual extraction or internal reasoning. By employing the MN (w/ GT-Map) setting, we provide models with perfect spatial knowledge. While the oracle map boosts global planning performance: increasing average $SR_\textbf{t}$ and $SR_{\textbf{p}}$ by 12.6\% and 8.1\%, which is remarkably limited. Even with ground-truth maps, the average $SR_\textbf{t}$ barely reaches 11.8\%, with a persistent $NE$ of 5.29m. This suggests that accurate spatial representation is a necessary but insufficient condition for mental navigation. The inability of MLLMs to reliably execute multi-step planning despite having full spatial priors proves that the bottleneck is rooted in a fundamental deficit of structured spatial reasoning mechanisms.

\subsection{The Horizon Collapse in Large-scale Planning.}

Analyzing performance across varying path lengths reveals a precipitous decay as spatiotemporal horizons expand. Under MN (w/ GT-Map), the average $SR_{\textbf{p}}$ drops by 2.1\% when transitioning from middle-range to long-range tasks. This performance collapse indicates that MLLMs largely rely on local heuristic strategies rather than coherent global planning. Much like traditional reactive VLN systems, these models fail to maintain spatial consistency when tasks require traversing multiple functional regions or retaining deep spatiotemporal dependencies, leading to a total failure of long-horizon internal simulation.




%% file: sec/4_learning_mental.tex
\section{Learning Mental Navigation with Structured Reasoning}
Motivated by these findings, we hypothesize that robust mental navigation requires two intertwined capabilities: explicit spatial representation (map construction) and structured reasoning over those representations (path planning). To this end, we propose NavMind, which reformulates spatial cognition as a decoupled, two-stage reasoning task. Given an egocentric memory video $V$ and query $q$, NavMind is mandated to perform: $(\hat{\mathcal{M}}, \hat{\mathcal{W}})=f_\theta(V, q)$. By compelling the model to first construct a hierarchical cognitive map $\hat{\mathcal{M}}$ before deriving a landmark-grounded reasoning chain $\hat{\mathcal{W}}$, we provide the necessary cognitive scaffold to overcome long-horizon spatial dependencies.

\begin{figure*}[t]
    \centering
    \includegraphics[width=1.0\textwidth]{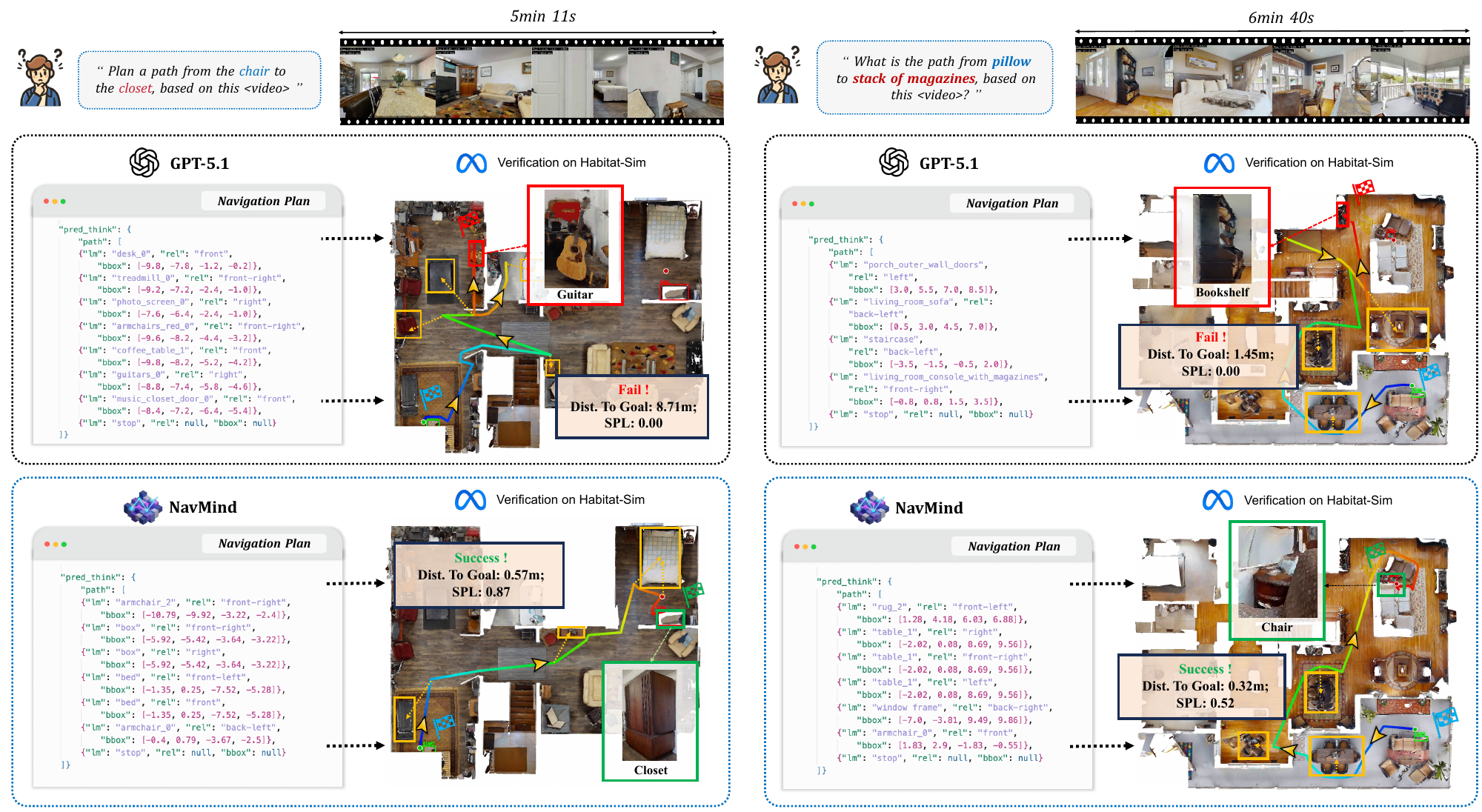}
    \caption{\textbf{Qualitative comparison.} We visualize the model-generated Navigation Plan (left) and and its Habitat-Sim verification (right). Blue flags denote the start point; red and green boxes indicate failed and successful endpoints. Yellow boxes and arrows mark landmark-grounded waypoints and predicted spatial relations. As illustrated, frontier MLLMs like GPT-5.1 exhibit spatial hallucinations and inaccurate reasoning, causing large plan deviations. In contrast, NavMind produces structured plans grounded in robust spatial representations, enabling efficient and successful execution.} 
    \label{fig:lidar}
    \vspace{-0.5cm}
\end{figure*}

\subsection{Cognition-Guided Progressive SFT}
To internalize these reasoning capabilities, we adopt a Cognition-Guided Progressive Training framework.Foundational Initialization. The pre-trained MLLMs (we adopt Qwen3-VL-8B in our setting) is first trained via Supervised Fine-Tuning (SFT) on full Video2Mind training set to map video sequences to ground-truth reasoning outputs $y^{\star}=({\mathcal{M}^\star}, {\mathcal{W}}^\star)$. The objective maximizes the likelihood of the structured sequence:$$\mathcal{L}_{\mathrm{SFT}}=\lambda_{\mathrm{map}} \mathcal{L}_{\mathrm{NLL}}\left(M^{\star} \mid V, q\right)+\lambda_{\text {think }} \mathcal{L}_{\mathrm{NLL}}\left(W^{\star} \mid V, q, M^{\star}\right)$$This stage establishes the fundamental mapping from visual observations to structured spatial knowledge.

We observe that standard datasets are often saturated with structurally simple episodes that offer weak learning signals, potentially encouraging the model to rely on trivial pattern matching. To resolve this, we introduce \textbf{Cognition-guided Rejection Sampling (CogRS)}, a difficulty-aware filtering strategy. Using the initial SFT model, we evaluate the perplexity of "decision-critical tokens": those representing key reasoning steps such as landmark selection and spatial relation inference. While low-perplexity samples are already mastered and extremely high-perplexity ones often contain noise, samples within a moderate perplexity interval $[\tau_{\min}, \tau_{\max}]$ provide the most informative supervision. By concentrating a second stage of progressive SFT on over 3,000 challenging trajectories, NavMind is compelled to internalize robust, multi-step spatial reasoning rather than memorizing templates.

\begin{figure*}[t]
    \centering
    \includegraphics[width=1.0\textwidth]{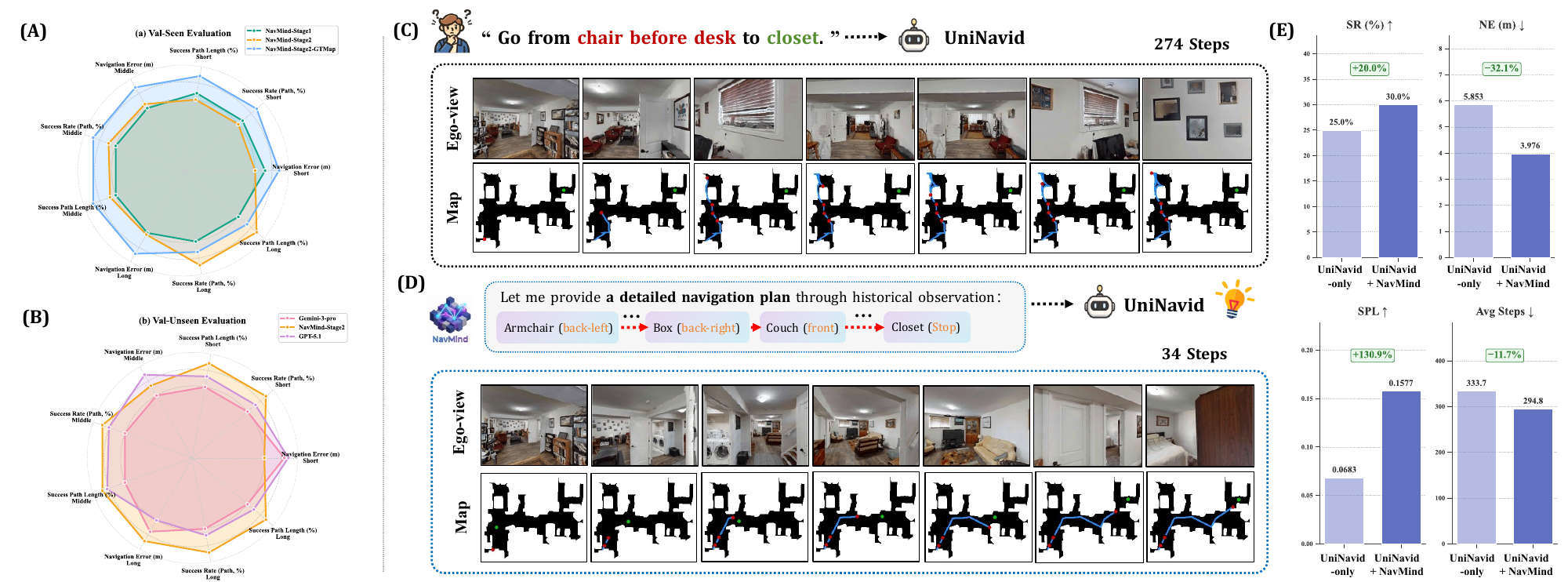}
    \caption{\textbf{Performance analysis and downstream integration of NavMind.}  \textbf{(A)} Comparison of NavMind's performance across different training stages. \textbf{(B)} Mental navigation performance in unseen scenes, demonstrating the model's generalization capability. \textbf{(C, D)} Comparison of recent VLN models using direct human instructions versus those incorporating NavMind’s fine-grained planning, which achieves significant improvements in navigation success rate and efficiency as quantitative in \textbf{(E)}.}
    \label{fig:vln}
    \vspace{-0.5cm}
\end{figure*}

\subsection{Experimental Results and Analysis}
For evaluation, we reconstruct navigable points in Habitat from the predicted landmarks and their spatial relations, simulate the navigation trajectory, and adjust the final position based on the predicted goal. $SR_\textbf{t}$ measures the success rate of reaching the final navigation point, while $SR_{\textbf{p}}$ measures the path success rate based on the final stopping location produced by the landmark-ground reasoning chain. This metric prevents models from achieving high scores by predicting only the final target while ignoring intermediate path nodes, providing a more reliable evaluation of navigation capability.

As shown in Tab.~\ref{table:benchmark_results_think}, NavMind significantly outperforms all baselines on the mental navigation task. Compared with the average baseline performance, NavMind improves $SR_\textbf{t}$/$SR_{\textbf{p}}$ by 43.2\%/34.2\%, reduces the navigation error by 3.33\,m, and increases route efficiency $SPL$ by 31.5\%. Moreover, as illustrated in Fig~\ref{fig:vln}.(A), NavMind shows substantial gains on longer tasks. In mid- and long-range navigation, $SR_{\textbf{p}}$ improves by 36.0\% / 30.5\%, while SPL increases by 33.5\% / 29.2\%. These results indicate that NavMind can perform long-horizon spatial reasoning based on the constructed cognitive map and enables more effective global navigation planning, demonstrating an initial form of mental navigation capability.

\noindent \textbf{Ablation Studies.} As shown in Fig~\ref{fig:vln}.(A), we further conduct ablation studies under the MN (w/ GT-Map) setting, where NavMind achieves additional improvements when provided with a more complete cognitive map. Notably, our hierarchical cognitive map better captures spatial relationships than flat grid-based representations while remaining more efficient than learning bounding boxes for all scene objects. Besides, the proposed CogRS mechanism yields larger improvements on medium- and long-range tasks. This suggests that CogRS effectively selects training samples that the model has not yet mastered, strengthening learning on challenging cases and improving training efficiency.

\noindent \textbf{Generalization in Unseen Environments.} To further evaluate the generalization ability of the proposed method, we conduct experiments in completely unseen environments. Specifically, we follow the same data generation pipeline used in Video2Mental and generate 350 high-quality navigation samples from the MP3D dataset as an unseen test set. Compared with HM3D, MP3D scenes exhibit more complex structures and more challenging navigation paths. As illustrated in Fig~\ref{fig:vln}.(B), when provided with the ground-truth cognitive map, NavMind demonstrates stronger spatial reasoning ability than GPT-5.1 and Gemini-3-Pro, while maintaining stable performance in complex environments. These results further highlight the potential of NavMind to construct global spatial navigation plans and generalize to unseen scenes.

\subsection{The Navigation Brain: Downstream VLN Integration}
The previous results indicate that NavMind already demonstrates preliminary mental navigation capability when executing navigation tasks. This suggests that it can serve as a reusable “navigation brain” to assist downstream navigation agents. In the earlier experiments, navigation paths were reconstructed in the Habitat-Sim environment by back-projecting predicted landmarks ($lm$) and spatial relations ($rel$). We further investigate whether NavMind can provide stable global planning signals when used as a planning module for VLN agents. To this end, we treat NavMind as the navigation brain, while Uni-NaVid~\cite{zhang2024uni} acts as the downstream policy agent in Habitat. For each predicted landmark ($lm$) and spatial relation ($rel$), Uni-NaVid generates the corresponding navigation actions and guides the agent along the planned route to reach the target.
As illustrated in Fig~\ref{fig:vln}.(C-D), under the same experimental setup, the VLN model fails to locate the target even after 274 actions, leaving a large distance from the goal. This highlights the limitation of existing VLN methods in long-horizon planning. In contrast, when NavMind is combined with Uni-NaVid, the agent reaches the target efficiently in only \textbf{34 actions,} significantly improving navigation efficiency. Furthermore, Fig~\ref{fig:vln}.(E) presents results across 20 different scenes, demonstrating that the global planning capability and structured cognitive map produced by NavMind can effectively assist VLN agents in navigating complex environments.




%% file: sec/5_conclusion.tex
\vspace{-0.2cm}
\section{Conclusion}
This work investigates whether MLLMs can perform \textbf{mental navigation} for long-horizon spatial reasoning and introduce \textbf{Video2Mental}, a benchmark designed for the mental navigation task that requires models to construct cognitive maps from egocentric videos and generate executable navigation plans verified in a simulator. Experiments reveal three insights: mental navigation does not naturally emerge from standard pre-training, spatial reasoning rather than perception is the primary bottleneck, and planning performance degrades as the navigation horizon increases.  To address these challenges, we propose \textbf{NavMind}, a Cognition-Guided progressive supervised fine-tuning framework that learns structured spatial reasoning via cognitive maps and landmark-grounded planning, improving navigation performance and enabling reusable global planning for VLN agents.